\documentclass[sigconf]{aamas}

\usepackage{balance} %
\usepackage{pgfplots}
\usepackage{multirow}
\usepackage{geometry}
\usepackage{tabularx}
\usepackage{booktabs}
\usepackage{ragged2e}

\setcopyright{ifaamas}
\acmConference[AAMAS '23]{Proc.\@ of the 22nd International Conference
on Autonomous Agents and Multiagent Systems (AAMAS 2023)}{May 29 -- June 2, 2023}
{London, United Kingdom}{A.~Ricci, W.~Yeoh, N.~Agmon, B.~An (eds.)}
\copyrightyear{2023}
\acmYear{2023}
\acmDOI{}
\acmPrice{}
\acmISBN{}

\usepackage{enumitem}
\usepackage{colortbl}
\usepackage{multicol}
\newcommand{\greyrule}{\arrayrulecolor{black!30}\midrule\arrayrulecolor{black}}

\acmSubmissionID{327}

\usepackage{pgfplots}
\pgfplotsset{width=8cm,compat=1.9}

\title[AAMAS-2023 Formatting Instructions]{Inferring Player Location in Sports Matches: Multi-Agent Spatial Imputation from Limited Observations}

\author{Gregory Everett}
\affiliation{%
  \institution{University of Southampton, UK}
  }
\email{gae1g17@soton.ac.uk}

\author{Ryan J. Beal}
\affiliation{%
  \institution{Sentient Sports, UK}
  }
\email{ryan.beal@sentientsports.com}

\author{Tim Matthews}
\affiliation{%
  \institution{Sentient Sports, UK}
 }
\email{tim.matthews@sentientsports.com}

\author{Joseph Early}
\affiliation{%
  \institution{University of Southampton, UK}
  }
\email{j.a.early@soton.ac.uk}

\author{Timothy J. Norman}
\affiliation{%
  \institution{University of Southampton, UK}
  }
\email{t.j.norman@soton.ac.uk}

\author{Sarvapali D. Ramchurn}
\affiliation{%
  \institution{University of Southampton, UK}
  }
\email{sdr1@soton.ac.uk}

\begin{abstract}
Understanding agent behaviour in Multi-Agent Systems (MAS) is an important problem in domains such as autonomous driving, disaster response, and sports analytics. Existing MAS problems typically use uniform timesteps with observations for all agents. In this work, we analyse the problem of agent location imputation, specifically posed in environments with non-uniform timesteps and limited agent observability ($\sim$95\% missing values). Our approach uses Long Short-Term Memory and Graph Neural Network components to learn temporal and inter-agent patterns to predict the location of all agents at every timestep. We apply this to the domain of football (soccer) by imputing the location of all players in a game from sparse event data (e.g., shots and passes). Our model estimates player locations to within $\sim$6.9m; a $\sim$62\% reduction in error from the best performing baseline. This approach facilitates downstream analysis tasks such as player physical metrics, player coverage, and team pitch control. Existing solutions to these tasks often require optical tracking data, which is expensive to obtain and only available to elite clubs. By imputing player locations from easy to obtain event data, we increase the accessibility of downstream tasks. 
\end{abstract}


\newcommand{\BibTeX}{\rm B\kern-.05em{\sc i\kern-.025em b}\kern-.08em\TeX}

\begin{document}

\pagestyle{fancy}
\fancyhead{}

\maketitle

\section{Introduction}

Many Multi-Agent System (MAS) applications involve agents interacting in both space and time to achieve their individual and collective goals. Predicting the behaviour of such agents is an important task in many areas such as path prediction for autonomous vehicles \cite{sriram2020smart}, predicting locations of civilians in disaster response scenarios using phone data or drone footage \cite{ramchurn2016disaster,wang2021pre}, and sports analytics \cite{omidshafiei2022multiagent}. However, situations may arise where the observability of the environment is limited. For example, an autonomous vehicle may have an obstructed view and lose sight of nearby pedestrians, or drone footage may have limited observability of an area when searching for injured civilians. In these situations, reasonable estimations of agent locations need to be made in order to improve the efficacy of response systems. In this work, we build a multi-agent time-series imputation model, named \textit{Agent Imputer}, which learns typical spatiotemporal interactions between agents to estimate the location of agents when system observability is limited. We apply this to the domain of Association Football
(soccer; referred to as football in the remainder of this work) by predicting the location of all players on a pitch when only observing the location and action (e.g., a pass or shot) of the ball-carrying agent.

Football represents an interesting domain for modelling teams as dynamic MAS, with each player having individual roles and behaviours. Whilst all players exhibit different behaviours, the system is centred around the ball. Therefore, in this work, we use data regarding the ball and ball-carrier to estimate all player locations when an event occurs. We build a model with Long Short-Term Memory (LSTM) and Graph Neural Network (GNN) components to learn the temporal and inter-agent relationships between players in a football game to make generalized predictions on unobserved player locations during a game.

Analytical data in football has many uses. These include modelling team tactics, calculating physical metrics (e.g., distance covered), and measuring pitch dominance of teams ~\cite{spearman_beyond_2018,link_real_2016,fernandez_decomposing_2019}. It can also be used by football clubs to perform player evaluation, scouting, and opposition analysis. In this work, we examine how sparse football event data of on-the-ball actions can be used to impute knowledge of player positions. This facilitates the desired downstream analysis tasks without requiring expensive tracking data.

\vspace{0.05cm}
\noindent Our contributions are as follows:
\vspace{-0.25cm}

\begin{enumerate}
  \item We propose a novel \textit{Agent Imputer} model that predicts agent locations when constrained to limited system observability. This model learns both temporal and inter-agent interactions using LSTM and GNN components to make estimations of agent positioning applied to football.
  \item We use event and tracking data from real-world football games to train and test our model, and find that it predicts agent positioning with a mean Euclidean distance error of $\sim$6.9m, outperforming numerous baselines by $\sim$62\%.
  \item We apply our model predictions to real-world downstream applications in the football domain, such as calculating estimations of player physical metrics, pitch dominance, and player coverage heatmaps.
\end{enumerate}
\vspace{-0.05cm}

\noindent The rest of this paper is structured as follows. Section \ref{background} discusses related literature and data in football, and Section \ref{playMM} formally defines the problem. Section \ref{pred-pmm} then introduces our novel \textit{Agent Imputer} model. Section \ref{emp-ev} contains our evaluation, and Section \ref{sec:applications} presents downstream applications. In Section \ref{discussion}, we discuss model outcomes and future work. Finally, Section \ref{conc} concludes.

\vspace{-0.2cm}
\section{Background} \label{background}
\vspace{-0.1cm}

In this section, we discuss the task of predicting agent behaviour in MAS (Section \ref{background-rw}), then cover existing uses of MAS in sports (Section \ref{background-pos-pred}), and finally motivate the need to impute player tracking by discussing available football data sources (Section \ref{background-data}).

\vspace{-0.2cm}
\subsection{Predicting Agent Spatiotemporal Behaviour} \label{background-rw}

There are a number of works in different MAS domains that predict the spatiotemporal behaviour of agents within dynamic multi-agent environments. These domains include collision avoidance for autonomous vehicles \cite{sriram2020smart,xie2021congestion}, pedestrian trajectory predictions in crowds \cite{ivanovic2019trajectron,alahi2016social,marchetti2020multiple}, location habits of human populations \cite{mcinerney2012improving,mcinerney2013modelling}, and sports analytics \cite{sun2019stochastic,le2017coordinated,omidshafiei2022multiagent}. We categorise existing work into two use-cases: predicting future behaviour, and imputing agent locations when behaviour is observed intermittently.

\vspace{-0.15cm}

\paragraph{Predicting Future Agent Behaviour} Given the previous behaviour of agents in a MAS, it can be useful to predict their future behaviour. A variety of approaches applied to different problems exist in the literature, such as LSTMs for estimating human trajectories within crowds \cite{alahi2016social, ivanovic2019trajectron}, variational recurrent neural networks for predicting future player trajectories in basketball \cite{sun2019stochastic}, and imitation learning to predict future trajectories of players in football \cite{le2017coordinated}.

\vspace{-0.15cm}

\paragraph{Imputing Agent Behaviour} In contrast to predicting future behaviour, imputation focuses on filling in missing information about agent behaviour using past and future observations. This is especially useful when behaviour is observed intermittently. An existing use case in football is the imputation of agent trajectories when players go out of camera view, which was achieved using GNNs and variational autoencoders \cite{omidshafiei2022multiagent}. Both future behaviour prediction and imputation can also be achieved with a single model \cite{qi2020imitative}. 

\vspace{-0.2cm}
\subsection{Multi-Agent Systems in Sports} \label{background-pos-pred}

Existing work in the sports domain has modelled teams as MAS to better understand team behaviour. This includes modelling networks of interactions between players to evaluate teamwork \cite{beal2020learning}, characterising team behaviour by mapping play sequences to game tactics \cite{lucey2012characterizing}, and classifying defensive actions to provide a framework for team tactical analysis \cite{raabe2022graph}. In contrast to modelling teams, several studies have modelled individual player behaviours in the context of MAS. This includes using imitation learning to identify multiple agent policies within a coordinated team structure \cite{le2017coordinated} and using LSTMs to predict trajectories of basketball players \cite{hauri2021multi}.

In contrast to existing work, instead of creating a model with a specific downstream sports analysis task in mind, we propose a model to impute missing player data, and then show how this facilitates a wide variety of applications that add value to those who can only access sparse sports data sources.

\subsection{Data in Football} \label{background-data}

The two main spatiotemporal data sources in football are tracking data and event data. We compare both types of data below.

\vspace{-0.1cm}

\paragraph{Tracking Data} The current gold standard in football analytics, optical tracking data records the location of every player and the ball $\sim$20 times per second  \cite{barris2008review}. This equates to ${\sim}108,000$ locations for a single player over the course of a game. Access to tracking data facilitates many industrial downstream use-cases for clubs to optimise their tactical setup or recruitment policies by analysing off-ball player movement and team performance \cite{fernandez_decomposing_2019, spearman_beyond_2018}. However, collecting this data requires expensive equipment to be installed within stadiums, and as such, it is currently mostly restricted to the top European leagues. Clubs without access to tracking data are unable to perform downstream analysis to the same extent as elite level clubs, leading to a wider gap in performance.

\vspace{-0.1cm}

\paragraph{Event Data} In contrast to high frequency tracking data, event data only records information on game events (e.g., passes and shots) along with the event location and the player involved \cite{pappalardo2019public}. A typical game involves ${\sim}3750$ events, or ${\sim}170$ events per player on average. This data is useful for on-ball statistics, but as it does not gather player information when they are off the ball, it cannot be readily used for many off-ball downstream tasks. However, this data is far cheaper and more easily attainable \textemdash{} there is an abundance of event data from many providers, covering 80+ global leagues.

\vspace{0.1cm}

\begin{figure}[!htb]
    \centering
    \includegraphics[width=\linewidth]{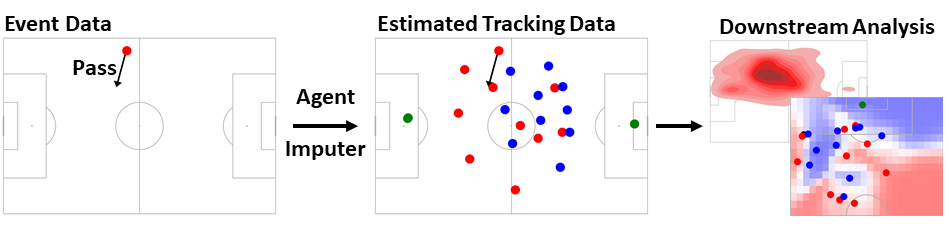}
    \caption{Comparison between event and tracking data for a single timeframe, and how our model estimates tracking data to facilitate downstream analysis tools.
    }
    \label{fig:eventTracking}
\end{figure}

\noindent In this paper, we propose a model that uses event data to impute off-ball player locations, and thus give snapshots of all player locations for all events during a game. This bridges the gap between event and tracking data, and facilitates the desired downstream analysis for clubs unable to access tracking data. Figure \ref{fig:eventTracking} gives a visual representation of the difference between event and tracking data, and how we use our model to estimate tracking data.

\vspace{-0.1cm}

\vspace{-0.25cm}
In contrast to the data used in existing work, event data in football leads to sparser agent observations with non-uniform timesteps. This occurs as the behaviour for an agent is only gathered when the player performs an on-the-ball action (e.g., pass, shot, tackle). This often leads to a cluster of agent events followed by an extended period with no information as the player is no longer on the ball. Furthermore, as only a single player can perform an on-the-ball event at a time, data is only gathered for a single agent at each timestep. As such, event data fails to capture important off ball information, such as attacking runs into space or collective defensive positioning, which is crucial when analysing team performance.

To our knowledge, no previous work has studied the problem we approach in this work: imputing agent behaviour within a MAS when information occurs at non-uniform timesteps and for a single agent at a time. In the next section, we formally define this problem.

\section{Problem Formulation} \label{playMM}

Our MAS is a collection of $N$ agents, $\textbf{A}=[a_{1},\ldots,a_{N}]$. In our target domain of football, $N=22$ (two teams of 11 on-field players). The time-series is a sequence of $T$ events \begin{math} \mathbf{E} = [\mathbf{e}_{1},\ldots,\mathbf{e}_{T}] \end{math}. Note this time-series is non-uniform, as each element $\mathbf{e_{t}} \in \mathbf{E}$ is recorded when an agent performs an on-ball action (e.g., a pass, dribble or shot), leading to varying gaps between each timestep.

For each event $\mathbf{e_{t}} \in \mathbf{E}$, we have a set of observations \begin{math}\Phi_{t} = [\phi_{t}^{1},\ldots,\phi_{t}^{N}] \end{math}, where $\phi_{t}^{n}$ is the observation of agent $n$ at timestep $t$. This gives a complete set of observations over time of \begin{math}\mathbf{\Phi} = [\Phi_{1},\ldots,\Phi_{T}] \end{math}. However, in our configuration, only one value in each $\Phi_{t} \in \mathbf{\Phi}$ is known, i.e., there are $N-1$ missing values for each $\Phi_{t}$, and $T(N-1)$ missing values in total. This occurs as we only make an observation of one agent at each timestep \textemdash{} in our target domain of football, this is the position of the on-the-ball agent. As the observed agent changes over time, we construct a one-hot encoded mask $\mathbf{M}$, where $\mathbf{M}_{t}^{n} = 1$ if agent $n$ is observed at timestep $t$, and is 0 otherwise. This ($T \times N$) binary matrix fully captures the information regarding known and unknown observations across the time-series problem.

In this configuration, the goal of an imputation model is to predict values for the unknown observations. Formally, for each $\mathbf{e_{t}} \in \mathbf{E}$, the model makes a prediction $\hat{\phi}_{t}^{n}$ for every $n \in [1,\ldots,N]$. This leads to a complete set of predicted observations \begin{math}\mathbf{\hat{\Phi}} = [\hat{\Phi}_{1},\ldots,\hat{\Phi}_{T}] \end{math}, where \begin{math}\hat{\Phi}_{t} = [\hat{\phi}_{t}^{1},\ldots,\hat{\phi}_{t}^{N}] \end{math}. Note, in this setup, the model is also predicting observations for cases that it has already observed (i.e., players on the ball). However, when we apply downstream analysis (Section \ref{sec:applications}), we instead use the actual locations for already observed agents. 

Relating the above to our target domain of football, the predicted observations $\mathbf{\hat{\Phi}}$ are the estimated locations of all players (both on-ball and off-ball) for every event  $\mathbf{e_{t}} \in \mathbf{E}$. The known observations are derived from the location of the events that occur. So, for event  $\mathbf{e_{t}} \in \mathbf{E}$ with an on-ball agent $a_n \in A$, the assigned observation $\phi_{t}^{n}$ is $e_{t}^{\mathtt{x,y}}$, where $e_{t}^{\mathtt{x,y}}$ is the $\mathtt{x,y}$ position at which $e_t$ occurred.

To summarise, the set of known observations $\mathbf{\Phi}$ (containing missing data) is a ($T \times N \times 2$) tensor, and the set of imputed observations $\mathbf{\hat{\Phi}}$ is also a ($T \times N \times 2$) tensor. Note, each observation is two-dimensional as we are using agent $\mathtt{x,y}$ positions \textemdash{} in other domains, these observations could be a different size. In the next section, we explain our model to solve this imputation problem.

\section{Agent Imputer Model} \label{pred-pmm}

The problem we describe in Section \ref{playMM} is highly complex as over 95\% of the values in $\boldsymbol{\Phi}$ are missing. We aim to extract useful information on agents' spatial and temporal behavioural patterns from event data, and apply it to a model which can consider how agents move over time and in relation to other agents. In this section, we outline the feature engineering (Section \ref{pred-pmm-pps}), model architecture (Section \ref{pred-pmm-xg}), and training process (Section \ref{sec:model_training}) used in our approach. For implementation details, see Appendix \ref{app:implementation}.

\subsection{Feature Engineering} \label{pred-pmm-pps}
The features available at a given timestep are strictly derived from on-the-ball event data, consisting of the event location, player on the ball, time at which the event occurred, and other information such as the current scoreline. This provides little to no information on the spatiotemporal context of an off-ball agent, such as when their location was last observed (i.e, when they were last on the ball). Therefore, we perform feature engineering on the event data to generate our own feature set which captures a more comprehensive view of each agent and the general movement of play. We create these features for each agent, so they are a combination of agent-specific and global features. For a timestep $t$ with event $\mathbf{e}_{t} \in \mathbf{E}$, and observed agent $a_n \in A$, we compute the following features:

\vspace{0.1cm}
\noindent\textit{Agent-Specific Features}
\vspace{-0.05cm}
\begin{itemize}[leftmargin=6pt]
\item[] \textbf{\texttt{\small{prevAgentTime,prevAgentX,prevAgentY}}}: time since the agent was last observed, and their location at that time. Formally, this uses the most recent previous timestep where the agent was on the ball: $t'$ s.t. $\mathbf{M}_{t'}^{n} = 1$ and $t' \leq t$, which can be found by iterating backwards in time from $t$. Note $t' = t$ if the agent is currently on the ball. Given $t'$, (\texttt{\small{prevAgentX}}, \texttt{\small{prevAgentY}}) = $e_{t'}^{\mathtt{x,y}}$ and \texttt{\small{prevAgentTime}} = $t-t'$. If the agent is yet to be observed, we impute these features with the values at the first timestep where they are observed.
\item[] \textbf{\texttt{\small {nextAgentTime,nextAgentX,nextAgentY}}}: time until the agent is next observed, and their location at that time. Similarly to the previous time and location features, this uses the soonest future timestep (including the current timestep) where the agent is on the ball: $t'$ s.t. $\mathbf{M}_{t'}^{n} = 1$ and $t' \geq t$, which can be found by iterating forwards in time from $t$. Given $t'$, (\texttt{\small{nextAgentX}}, \texttt{\small{nextAgentY}}) = $e_{t'}^{\mathtt{x,y}}$ and \texttt{\small{nextAgentTime}} = $t'-t$. If the agent has no future observations, we impute these features using the last time they are observed.
\item[] \textbf{\texttt{\small{avAgentX,avAgentY}}}: mean location of the agent across the entire game, i.e., the mean position of all events where $\mathbf{M}^n = 1$.
\item[] \textbf{\texttt{\small{agentRole}}}: the agent's role in the team (e.g., central defender or goalkeeper). Different data providers label roles differently \textemdash{} for our data source, there are 16 possible roles, see Appendix \ref{app:positions}.

\item[] \textbf{\texttt{\small{agentSide}}}: binary indicator of whether an agent is on the same team as the current ball-carrying agent. 
\item[] \textbf{\texttt{\small{agentObserved}}}: binary indicator of whether the agent is the one performing the current on-ball action.
\item[] \textbf{\texttt{\small{goalDiff}}}: the difference in score between the teams (number of goals for this agent's team - number of goals for the other team).
\end{itemize}

\vspace{0.05cm}
\noindent\textit{Global Features}
\vspace{-0.05cm}
\begin{itemize}[leftmargin=6pt]
\item[] \textbf{\texttt{\small{eventX,eventY}}}: location of the current event (i.e. $e_{t}^{\mathtt{x,y}}$).
\item[] \textbf{\texttt{\small{eventType}}}: the type of the current event (e.g., a pass or shot). For more details, see Appendix \ref{app:dataset}.
\end{itemize}

\noindent These 15 features, for each agent at each timestep, represent a transformation of the original event data. This transformation means the data is more readily usable by machine learning methods. Note that agent positions are relative to the agent's own goal-line, irrespective of direction of play (as opposed to absolute position on the pitch). 
In the next subsection, we detail our \textit{Agent Imputer} model.

\subsection{Model Architecture} \label{pred-pmm-xg}

\begin{figure*} 
  \includegraphics[width=0.98\textwidth]{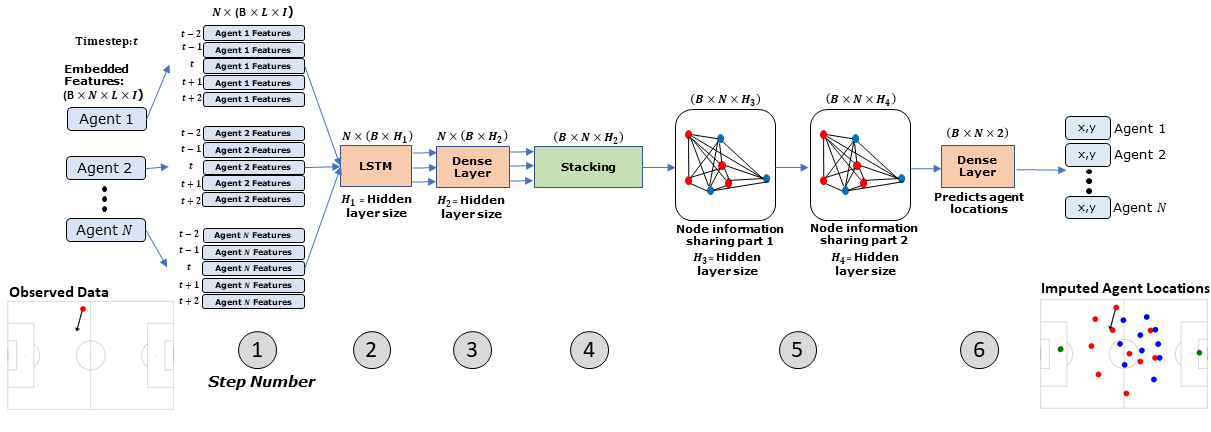}
  \vspace{-0.6cm}
  \caption{Agent Imputer model architecture. All agent location estimations for an event are made using agent feature information described in Section \ref{pred-pmm-pps}, with batch size $B$, number of agents $N$, sequence length $L$, and number of features $I$.}
  \label{fig:model}
  \vspace{-0.3cm}
\end{figure*}

Our \textit{Agent Imputer} model estimates the set of agent locations $\hat{\Phi}$ using the engineered features from the set of events $\mathbf{E}$. The goal of this model is to capture agent movement over time, the relationship between on-ball events and off-ball agent locations, and how agents interact as a team within the MAS. In football, this is often defined as team structure. We build an architecture which learns across time and agents within the MAS. Below, we explain this model using a step-by-step process, with corresponding elements of the architecture marked in Figure \ref{fig:model}.

\subsubsection{\textbf{Step 1: Input Formatting}}
To capture the ball and agent locations across time, we input a window of event data. We extract the feature set defined in Section \ref{pred-pmm-pps} for a sequence of $L=5$ events. This sequence is centred around a particular timestep $t$, i.e., we compute the features for timesteps $\{t-2,\ldots,t+2\}$.\footnote{If timesteps before or after the current timestep don't exist due to being at the start or end of a game, the data for the current timestep is used instead.} 
Each categorical feature (\texttt{\small{agentRole}}, \texttt{\small{agentSide}}, \texttt{\small{agentObserved}}, \texttt{\small{goalDiff}}, \texttt{\small{eventType}}) is converted to an embedding (see Appendix \ref{app:encoding}
for details). Our imputation model takes all the agent feature sets as a single simultaneous input, such that the model learns the spatial structure of the whole system (as opposed to processing each agent independently). Therefore, the input data has shape ($B \times N \times L \times I$), where $B$ is the batch size, $N$ is the number of agents, $L$ is the sequence length, and $I$ is the post-embedding dimensionality of our feature set including both the numerical features and embedded categorical features. In this work, $I = 24$ and $B=128$.

\subsubsection{\textbf{Step 2: LSTM Component}}

Each agent is independently passed into a shared bidirectional LSTM component, i.e., the input data is split into $N$ segments of size ($B \times L \times I$). The LSTM is then able to learn the temporal relationship between the engineered features and agent location. Importantly, the LSTM is shared across all agents. This overcomes issues with the sparsity of agent observations, as the LSTM is able to learn agent positioning through common movement patterns for agents with similar roles. Due to the irregular time intervals between timesteps, it is important to augment the LSTM architecture to deal with non-uniformity. Therefore, we use a Time-Aware LSTM \cite{baytas2017patient} which adjusts cell memory to alter the discount rate of previous or future actions in the sequence based on the difference in time from the current event. In this work, an LSTM with a single hidden layer of size $H_{1} = 100$ is used.

\subsubsection{\textbf{Step 3: Dense Layer with ReLU Activation}}

Outputs from the LSTM model for each agent at the current timestep $t$ (i.e., the middle of the input sequence) are extracted. This output is passed through a dense layer with a ReLU activation function to give an output of size $H_{2} = 50$, resulting in a latent representation of the time-aggregated features for each agent.

\vspace{-0.1cm}
\subsubsection{\textbf{Step 4: Stacked Agent Embeddings}}

The temporally aggregated latent representations are stacked together, giving an output tensor of size ($B \times N \times H_{2}$). This data is now a suitable input for a graph neural network (GNN), with each tensor row representing agent information that become node features for the GNN.

\vspace{-0.1cm}
\subsubsection{\textbf{Step 5: GNN Component}}

The LSTM component of the network models the temporal aspect of agent behaviour. We now look to also model the inter-agent relationships within the MAS, and how these impact the behaviour of each agent. We construct a fully connected graph with the temporally-aggregated agent representations as node features. The GNN uses this graph structure to allow information sharing across all agents. The GNN architecture consists of two message passing layers with feature sizes of $H_{3} = 64$ and $H_{4} = 32$, using the SAGEConv operator \cite{hamilton2017inductive}. This updates node features by using a mean aggregation scheme to learn information about agent neighbourhoods (i.e., agent interactions).

\vspace{-0.1cm}
\subsubsection{\textbf{Step 6: Dense Layer with ReLU Activation}}

The final output of the GNN component is of size ($B \times N \times H_{4}$), i.e., latent representations of size 32 for each agent, which captures both temporal and inter-agent interaction information. These final representations are then passed through a single dense layer with a ReLU activation function to make the final position (\texttt{x,y}) predictions for each agent. Therefore, the final model output is of size ($B \times N \times2$).

\vspace{-0.1cm}
\subsection{Model Training} \label{sec:model_training}

We use tracking data as target variables during training, and as ground truth labels to test the predictive accuracy of our \textit{Agent Imputer} model. We use the mean Euclidean distance between the predicted positions and true positions as the loss function to train the model. The model was trained for 150 epochs with a batch size of 128. The AdamW \cite{loshchilov2018decoupled} optimiser was used with an initial learning rate of 0.002. These hyper-parameters and the exact model architecture (e.g., hidden layer sizes) were found through trial and error, i.e., no formal hyper-parameter tuning was carried out.

\begin{table*}[htb!]
\caption{Predictive performance of models averaged across five folds, along with 95\% confidence intervals. We give performance in the X, Y, and Euclidean (XY) directions. All errors are in metres. Bold results indicate the best performance (lowest error).}
\vspace{-0.3cm}
\label{tab:pred-performance}
\begin{tabular}{lllllll}
\toprule
 &  \multicolumn{3}{c}{Train}  &  \multicolumn{3}{c}{Test} \\
\cmidrule(lr){2-4}\cmidrule(lr){5-7}
Models & X Error & Y Error & XY Error & X Error & Y Error & XY Error \\
\midrule
Baseline 1 & 14.20 $\pm$ 0.02  & 9.69 $\pm$ 0.02 & 18.81 $\pm$ 0.05 & 14.22 $\pm$ 0.24 & 9.70 $\pm$ 0.14 & 18.82 $\pm$ 0.22\\
Baseline 2 & 14.10 $\pm$ 0.02  & 9.79 $\pm$ 0.02 & 18.91 $\pm$ 0.05 & 14.04 $\pm$ 0.48 & 9.78 $\pm$ 0.17 & 18.80 $\pm$ 0.50 \\
Baseline 3 & 13.37 $\pm$ 0.05  & 9.63 $\pm$ 0.01 & 18.15 $\pm$ 0.05 & 13.27 $\pm$ 0.44 & 9.57 $\pm$ 0.14 & 18.01 $\pm$ 0.46\\
\greyrule
XGBoost & 5.88 $\pm$ 0.01 & 5.14 $\pm$ 0.02 & 8.67 $\pm$ 0.02 & 6.42 $\pm$ 0.35 & 5.50 $\pm$ 0.09 & 9.26 $\pm$ 0.26\\
Time-Aware LSTM & 4.46 $\pm$ 0.02 & 4.26 $\pm$ 0.02 & 6.89 $\pm$ 0.02 & 4.48 $\pm$ 0.08 & 4.49 $\pm$ 0.1 & 7.09 $\pm$ 0.06\\
GNN & 5.28 $\pm$ 0.05 & 5.08 $\pm$ 0.1 & 8.18 $\pm$ 0.09 & 5.42 $\pm$ 0.35 & 5.13 $\pm$ 0.14 & 8.32 $\pm$ 0.25\\
\greyrule
\textit{Agent Imputer} & \textbf{4.06 $\pm$ 0.02} & \textbf{4.12 $\pm$ 0.02} & \textbf{6.47 $\pm$ 0.02} & \textbf{4.29 $\pm$ 0.09} & \textbf{4.41 $\pm$ 0.11} & \textbf{6.88 $\pm$ 0.1}\\
\bottomrule
\end{tabular}
\vspace{-0.1cm}
\end{table*}

\section{Empirical Evaluation} \label{emp-ev}

In this section, we describe the data used to train and evaluate our model (Section \ref{sec:datasets}), introduce the baselines we compare to (Section \ref{sec:baselines}), and give our results (Section \ref{sec:results}). 

\vspace{-0.1cm}
\subsection{Datasets} \label{sec:datasets}

We train and evaluate our model using 34 games of event and tracking football data collected from K League 1\footnote{The top men's professional football division in South Korea} and supplied to us by Bepro Group Ltd. These are gold standard industry datasets that allow us to rigorously evaluate our approach. Each game provides event sequences, which we use as model input, and tracking data, which we use as training targets and for evaluation. In total, there are ${\sim}$64,000 events and ${\sim}$1.4 million tracking locations. We use a 31/3 (${\sim}$91.2\%/8.8\%) train/test split and five-fold cross validation to evaluate our model. All geometric data is scaled to a standard football field size of 105x68 metres. We provide further details on the dataset in Appendix \ref{app:dataset}.

\vspace{-0.1cm}
\subsection{Baselines} \label{sec:baselines}

We use a number of baselines to evaluate whether our model has improved performance compared to other imputation methods. We first propose three na\"ive imputation baselines. Baseline 1 simply predicts agent location using the average on-ball location of the agent during the match. Baseline 2 predicts the centroid between the last and next observed location of the agent, and Baseline 3 predicts a time-scaled position on the straight-line trajectory between the last and next observed location of the agent. We also compare our \textit{Agent Imputer} model to other machine learning models with the same input feature set, including an XGBoost regression model, a Time-Aware LSTM model (i.e., the same model as in Figure \ref{fig:model} without step 5), and a GNN model (i.e., steps 4, 5 and 6 of Figure \ref{fig:model}, using the middle of the original input sequences as input features). These last two baselines allow us to compare how the constituent parts of our \textit{Agent Imputer} model contribute to performance.

\vspace{-0.1cm}
\subsection{Results} \label{sec:results}

Four experiments are used to evaluate the performance of our \textit{Agent Imputer} model and relevant baselines. We evaluate player position prediction (Section \ref{sec:results_overall}), performance over time (Section \ref{sec:results_time}), performance over different positions (Section \ref{sec:results_role}), and how performance varies with observation (Section \ref{sec:results_last_observed}).

\subsubsection{Position Prediction} \label{sec:results_overall}

We evaluate the predictive performance of the model and baselines in Table \ref{tab:pred-performance}. Note the X direction is along the length of the pitch (105m) and the Y direction is along the width of the pitch (68m), and that we do not normalise the error in these different directions by pitch size. We find that the \textit{Agent Imputer} model predicts agent location with highest accuracy (lowest distance error). We also find that the Time-Aware LSTM model outperforms the GNN model, suggesting the essential part of our \textit{Agent Imputer} is the LSTM component. However, we still see an improvement in performance by including the GNN component to the \textit{Agent Imputer} model, demonstrating the value of modelling agent interactions. We note a 61.8\% decrease in error using the \textit{Agent Imputer} compared to the best performing na\"ive baseline. We also find the \textit{Agent Imputer} model has roughly equal error in both X and Y directions.

\subsubsection{Predictability over Time} \label{sec:results_time}

\begin{figure}[htb!]
    \centering
    \includegraphics[width=\linewidth]{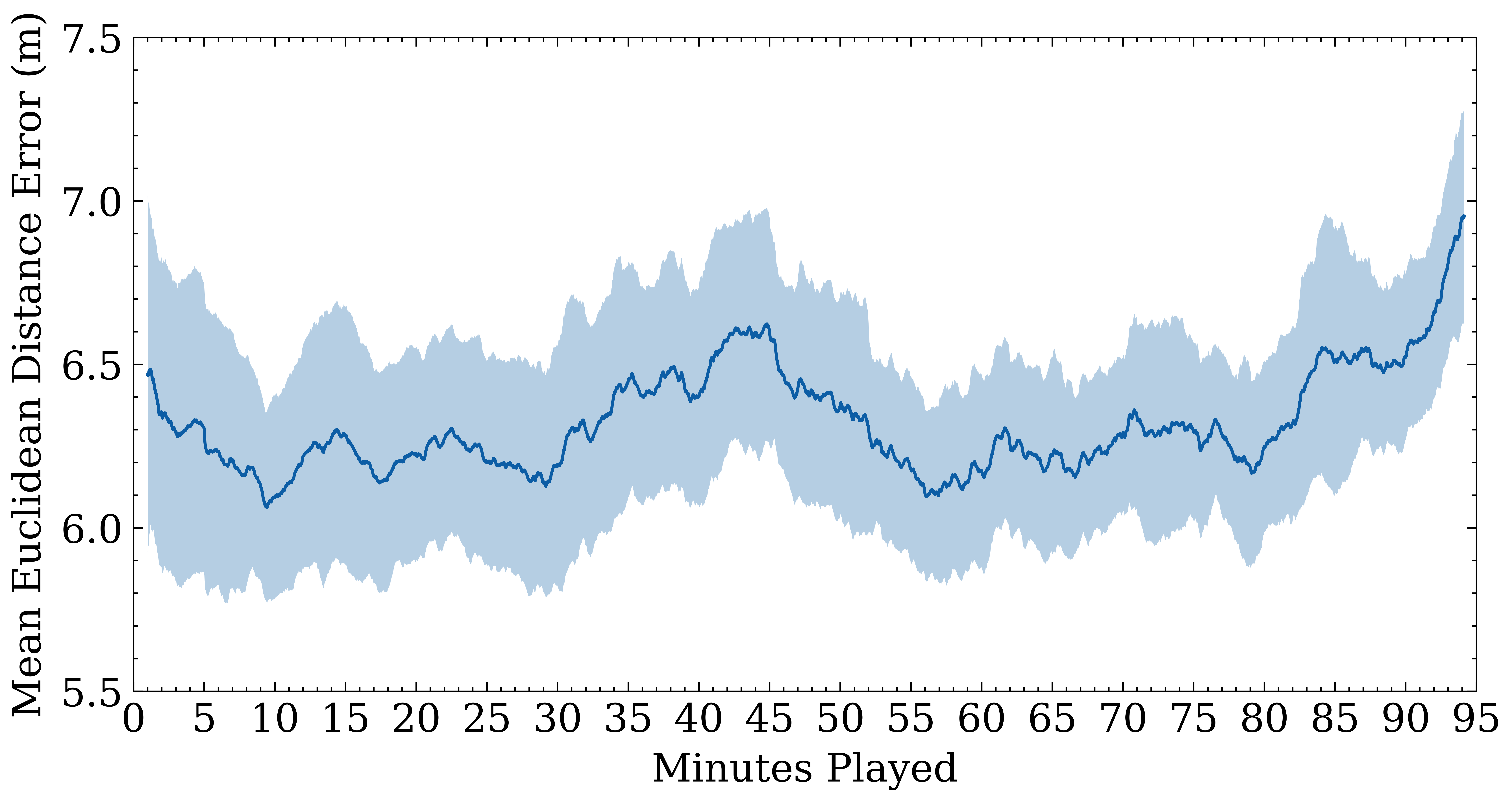}
    \vspace{-0.6cm}
    \caption{Predictive performance of the \textit{Agent Imputer} model across time using a rolling average of Euclidean error (5 minute windows with a 1 second step size) on the combined training and testing data (all 34 games). Shading shows the standard deviation across means for each game.}
    \label{fig:error_over_time}
\end{figure}

We evaluate predictive accuracy of the model across different periods of a match. In football, the game has two 45 minute halves, so we evaluate how model accuracy changes during these periods. We compute this using a rolling average of model error; see Figure \ref{fig:error_over_time}. Additional time at the end of halves is accounted for \textemdash{} first half added time is merged with early second half predictions, and values beyond 90 minutes facilitate second half added time.

\vspace{-0.2cm}
\vspace{-0.3cm}

\begin{figure*}[t!]
    \centering
    \includegraphics[width=\linewidth]{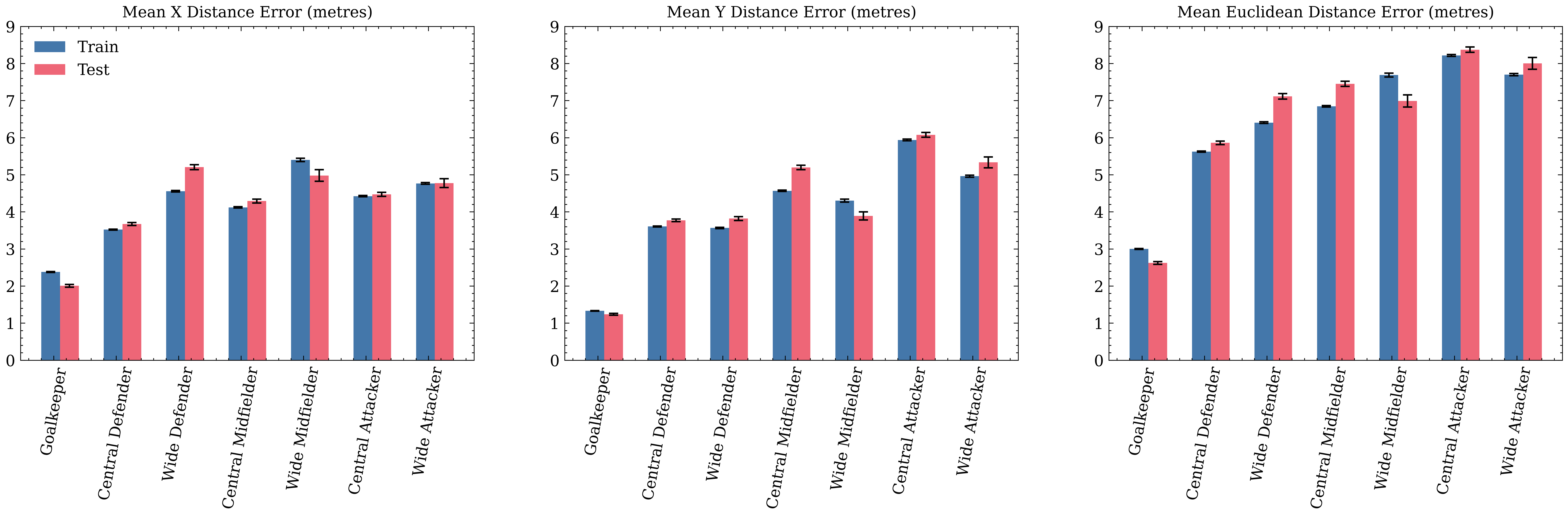}
    \vspace{-0.6cm}
    \caption{Mean distance error of player location estimations for each agent role using the \textit{Agent Imputer} model.}
    \vspace{-0.2cm}
    \label{fig:dist_per_pos}
\end{figure*}

From these results, we suggest that the most unpredictable moments of a game are at the end of both halves. Due to the error margin, we perform t-tests to investigate the hypothesis that the mean of the underlying sample distribution in the middle of a half is lower than the underlying distribution at the end of a half. When doing this for the the first half (all events between 20-25 minutes compared to all events between 42.5-47.5 minutes), 
this is found to be significant (p<0.01). The difference in the second half (comparing all events between 65-70 minutes with all events between 87.5-92.5 minutes), is also found to be significant (p<0.01). 

This supports our theory that the game is more unpredictable at the end of halves, which corroborates the popular intuition that players get tired and teams become less structured at the end of halves. Furthermore, teams are looking to get back into a game or defend a lead, which leads to more chaotic and unpredictable periods. This highlights match phases at which team structure varies, and this type of analysis may also uncover situations where players are moving in `unusual' ways. This type of analysis could be extended to identify links between unusual movement patterns and team performance, and used as a possible tool to proactively identify performance change during a match.

\vspace{-0.2cm}

\subsubsection{Predictability over Roles} \label{sec:results_role}

We evaluate the performance of our \textit{Agent Imputer} model across agent roles in Figure \ref{fig:dist_per_pos}. This gives insight into how an agent's goals and responsibilities within a team affects the predictability of their behaviour. For simplicity, in this experiment and downstream applications (Section \ref{sec:applications}), we group roles into wide and central positions, reducing the total number of positions from 16 to 7 (see Appendix \ref{app:positions}). 
Goalkeepers are expectedly the most predictable role, as their range of movement is usually limited to their own box. We also find that defenders are more predictable than attackers, highlighting that defensive agents behave in a more structured way than attacking agents. 

Interesting findings can also be drawn from the X and Y distance errors. A general trend is that the model finds it harder to predict the X location of wide outfield players in comparison to central outfield players (an average of 4.99m vs 4.15m), which is expected as wide players cover a lot of ground along the wings of the pitch. However, the model generally predicts the Y location of wide players better than central players (4.35m vs 5.02m), which is also expected as wider players usually stick to a single side of the pitch. These comparisons help identify where the model could be improved, and could be used to help convey uncertainty in downstream analysis.

\subsubsection{Predictive Performance over Observation Offset} \label{sec:results_last_observed}

\begin{figure}[!htb]
    \centering
    \includegraphics[width=\linewidth]{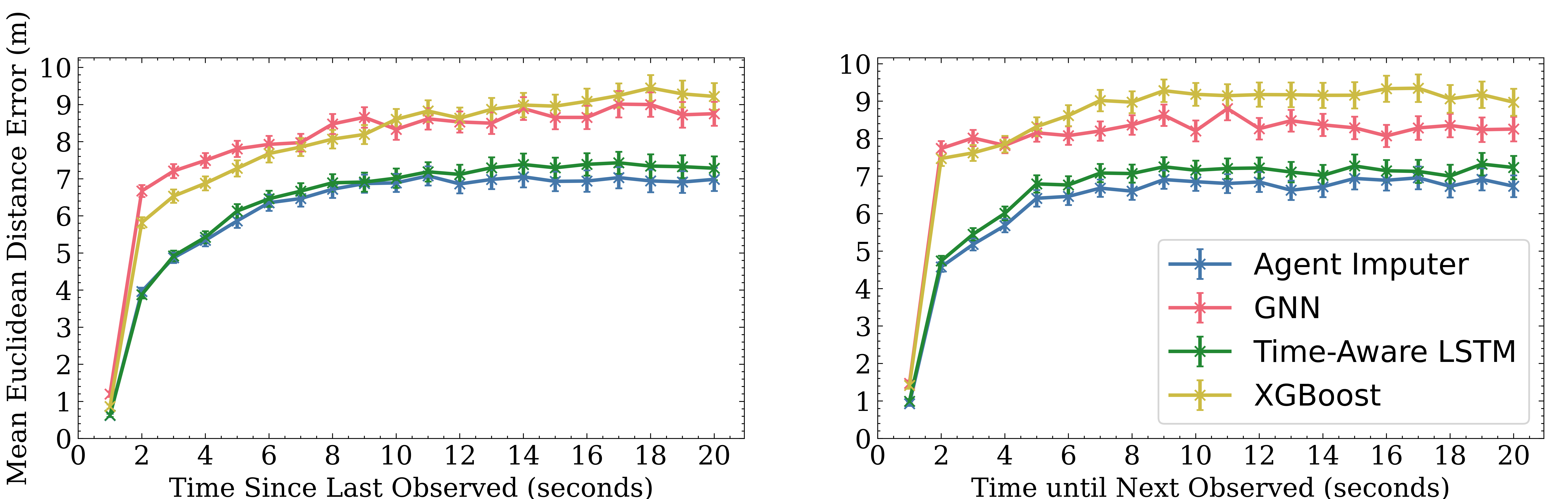}
    \vspace{-0.6cm}
    \caption{Model performance compared to the last (left) and next (right) time the agent was observed. We use one second intervals with a one second window, e.g., at 2s, we evaluate for all events between 1-2 seconds.}
    \vspace{-0.3cm}
    \label{fig:error_since_observed}
\end{figure}

In Figure \ref{fig:error_since_observed}, we evaluate the predictive performance of our proposed models with respect to ``time since last observed'' \textemdash{} the amount of time that has transpired since a particular agent was last involved in an event. We also do the same for ``time until next observed'' \textemdash{} the amount of time until the agent is next involved in an event. We find that the performance of the XGBoost and GNN model rapidly decrease as time increases. In comparison, the \textit{Agent Imputer} and Time-Aware LSTM models show a slower decay in performance, as well as a lower plateau. This demonstrates the necessity of modelling temporal aspects of predictions. However, the time-aware models still decrease in performance as time increases. Similarly to the error by player role (Section \ref{sec:results_role}), this may be useful in measuring the uncertainty of model predictions, i.e., there is greater certainty in predictions for agents that have been on the ball more recently.

\section{Model Applications} \label{sec:applications}

In this section, we demonstrate downstream analysis that can be performed using outputs from our \textit{Agent Imputer} model. In the football domain, these types of analyses are often implemented by elite level clubs using tracking data. However, we demonstrate that lower league clubs with limited resources could use our model to perform similar processes to the elite level clubs without requiring expensive and hard-to-obtain tracking data. We apply our model to analysing player physical metrics (Section \ref{sec:analysis_distance}), pitch control (Section \ref{sec:analysis_control}), and positional heatmaps (Section \ref{sec:analysis_heatmap}).

\subsection{Player Physical Metrics} \label{sec:analysis_distance}

We show how predicted player locations from our \textit{Agent Imputer} model can be used to estimate player physical metrics. We focus on distance covered, which is calculated by summing the Euclidean distances between a player's predicted locations throughout a match. We initially found that the predicted distance covered is consistently higher than the actual distance covered, showing an average overestimate of 11.5\% for \textit{Agent Imputer}, 16.8\% for the Time-Aware LSTM, 31.7\% for the GNN, and 29.7\% for XGBoost.

This overestimation bias can be attributed to the large number of events that occur soon after another event \textemdash{} 29.9\% of events happen within one second of the previous event. Typically, these are duels between a player on each team, or a player receiving the ball and then quickly passing. In these instances, the players' predicted positions shift markedly, i.e., moving faster than the maximum player speed. This implies the models are not considering realistic player trajectories between events which occur in quick succession. 
Fixing this model issue is future work. For now, we use a post-processing step that combines events which occur within a second of each other, using the model output for the initial event as the player location when grouping. This leads to more accurate distance covered estimations.
Results for the predicted distance covered averaged across different player roles are shown in Table \ref{tab:dist_covered}.\footnote{Distance is normalized as if all players play for 90 minutes (i.e., accounting for extra time and substitutions). Players that played for less than 20 minutes were excluded from these results. Results averaged across all games for players in a certain role.}

\begin{table}[!h]
\caption{Player distance covered results. Distance is averaged over the test data and given in kilometres. Absolute error is averaged over each game for each player role.}
  \label{tab:dist_covered}
  \begin{tabular}{llll}
    \toprule
    Role & Pred. Dist. & True Dist. & Abs. \% Error \\
    \midrule
    Goalkeeper & 3.23 & 3.12 & 4.05 $\pm$ 2.52\\
    \greyrule
    Central Defender & 8.26 & 8.42 &4.31 $\pm$ 1.38\\
    Wide Defender & 8.97 & 9.08&4.99 $\pm$ 1.89\\
    \greyrule
    Central Midfielder & 9.37 & 9.48&3.66 $\pm$ 1.26 \\
    Wide Midfielder & 8.73 & 8.97&2.79 $\pm$ 0.76\\
    \greyrule
    Central Attacker & 8.34 & 8.35&6.03 $\pm$ 2.32\\
    Wide Attacker & 8.26 & 8.70&5.33 $\pm$ 0.76\\
   \bottomrule
  \end{tabular}
\end{table}

\subsection{Pitch Control Analysis}  \label{sec:analysis_control}

To test a model application which considers agent locations relative to each other within the system, we calculate pitch control of teams.
Pitch control \cite{spearman2017physics,spearman_beyond_2018,fernandez_decomposing_2019} is a popular downstream analysis tool within football analytics that uses player locations and trajectories to compute the area in which a team dominates. It splits the pitch into a grid of zones and computes the probability of the attacking team controlling the ball if it arrived in that zone. For example, if a defending player is closer to the zone than the nearest attacker, it is unlikely the attacking team will control the ball in that zone. We perform similar analysis to that used in \cite{omidshafiei2022multiagent} \textemdash{} we compute pitch control using \textit{Agent Imputer} outputs and compare with the same computed from actual player locations in tracking data. Table \ref{tab:pitch_control} shows the mean absolute error of pitch control for each model, and shows that the \textit{Agent Imputer} performs best.

\begin{table}[h!]
\caption{Pitch control performance compared to the ground truth. Calculated over test data.}
  \label{tab:pitch_control}
  \begin{tabular}{ll}
    \toprule
    Model& Mean Average Error\\
    \midrule
    Baseline 1 & 0.272 $\pm$ 0.002 \\
    Baseline 2 & 0.304 $\pm$ 0.002 \\
    Baseline 3 & 0.305 $\pm$ 0.002 \\
    \greyrule
    XGBoost Regression & 0.150  $\pm$ 0.001\\
    GNN & 0.149 $\pm$ 0.001 \\
    Time-Aware LSTM & 0.135 $\pm$ 0.001\\
    \greyrule
    \textit{Agent Imputer} & \textbf{0.130 $\pm$ 0.001}\\
  \bottomrule
\end{tabular}
\end{table}

\begin{figure}[htp]
    \centering
    \includegraphics[width=\linewidth]{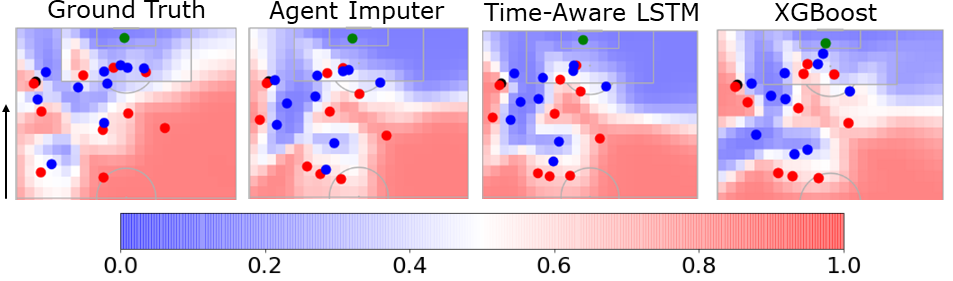}
    \caption{Pitch control diagrams comparing three models with ground truth. Red regions are in control of the attacking team, and blue regions are in control of the defending team. We only show the attacking half as the remainder of the pitch is entirely under the attacking team's control (red). Arrow refers to attack direction.} 
    \label{fig:pc}
\end{figure}

We provide example of pitch control outputs in Figure \ref{fig:pc}. These visualisations can be used by clubs to review their team's dominance in particular pitch areas in different game scenarios. This plot also highlights the differences in predicted player locations between the different models. It can be seen that, for the defending team, players are further apart and a structured horizontal defensive line is shown in the \textit{Agent Imputer} predictions. Whereas for the other models, some defenders can be seen to be stacked vertically - a very unlikely prediction given the usual defensive setup and strategy employed by football teams. This suggests that the GNN module is correctly learning how agents spatially interact.

\vspace{-0.1cm}

\subsection{Player Heatmap Analysis}  \label{sec:analysis_heatmap}

\begin{figure}[htp]
    \centering
    \includegraphics[width=\linewidth]{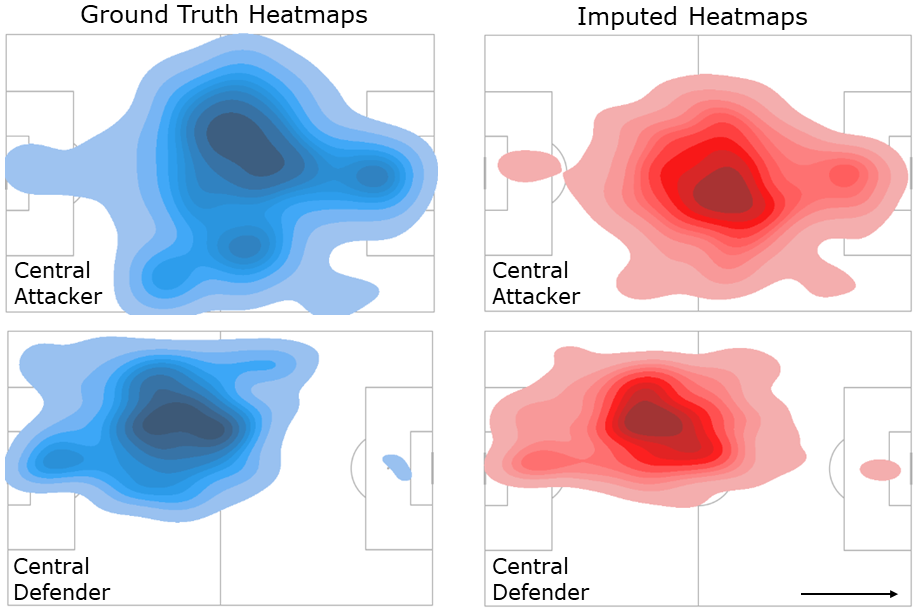}
    \caption{Example \textit{Agent Imputer} heatmaps compared with ground truth heatmaps. Darkest regions indicate most frequently occupied areas. Players are playing from left to right. We give further plots in Appendix \ref{app:heatmaps}.
    }
    \label{fig:heatmap}
    \vspace{-0.5cm}
\end{figure}

In this section, we use predicted player locations from the \textit{Agent Imputer} model to generate player heatmaps over an entire game. This provides coaches with useful information on the areas most frequently covered by their players. Note, while it is possible to generate heatmaps from event data, due to the nature of the data, they will be missing $\sim$95\% of player positions. Figure \ref{fig:heatmap} presents some comparisons between heatmaps of a central attacker and central defender using imputed positions.

The imputed heatmaps show many similarities to the ground truth heatmaps. Note the model has picked up interesting features, such as the small distinct region for the central defender near the opponent's goal. This most likely stems from set-pieces (such as a corner), demonstrating the \textit{Agent Imputer} has learnt features of the game state and how this influences player position. Furthermore, the central defender heatmap is offset to the left-hand wing of the pitch. In comparison, the central attacker heatmap which is more central. This shows the model has learnt role-specific information \textemdash{} typically there are two or three central defenders, allowing them to focus on particular sides of the pitch, whereas a central attacker is likely more free to roam in an unstructured manner.

\vspace{-0.1cm}
\section{Discussion} \label{discussion}
In this section, we further discuss our findings (Section \ref{sec:discussion_analysis}) and highlight limitations and areas of future work (Section \ref{sec:discussion_future}).

\vspace{-0.1cm}
\subsection{Analysis of Results} \label{sec:discussion_analysis}

As shown by our results (Section \ref{sec:results}), our model is able to impute agent locations using very sparse data, and do so with greater accuracy than na\"ive baselines and other machine learning models. The increase in performance compared to XGBoost suggests that the temporal and inter-agent dynamics captured by the LSTM and GNN components offer value in learning agent location.

Our \textit{Agent Imputer} model also enables easier access to insightful use-cases in football, such as player physical metrics, player heatmaps, and team pitch dominance analysis (see Section \ref{sec:applications}). These use-cases could be implemented by lower league clubs with fewer resources to perform similar player evaluation, scouting, and game analysis processes to elite level clubs. This would help close the performance gap between these teams.

As an example, the player heatmaps are shown to be strongly correlated to the ground truth heatmaps exhibited in-game. This can provide value to an opposition analytics scout who could monitor an opponent and get their players to exploit areas that are frequently left uncovered by the opposition team. It may also allow scouts to find players who occupy dangerous space more often.

The predictive ability of our model demonstrates that positioning in football is somewhat predictable. This is likely due to tactical positioning, where teams will be set up to defend and attack in consistent shapes. Work that considers the game theoretic implications of team tactics may introduce further insight into the predictability of teams and how this varies over time \cite{beal2020optimising,beal2021optimising}. Furthermore, players in certain roles will be given instructions on where to position themselves in certain patterns of play. These patterns can be learned by our model to predict future occurrences of such positioning.

\vspace{-0.1cm}
\subsection{Future Work and Limitations} \label{sec:discussion_future}

Our \textit{Agent Imputer} approach could also be applied to other MAS with limited observability, such as searching for injured civilians in disaster response systems or predicting human movement from phone location data \cite{mcinerney2012improving,mcinerney2013modelling}. Change of domain will inevitably lead to some differences in problem structure, such as the bounds of the environment or the sparsity of the dataset. These differences must be considered when applying our method to other domains. Another research direction would be to make the model probabilistic so that uncertainty in agent locations can be quantified.

Our model performs well at estimating location. However, as discussed in Section \ref{sec:results_time}, extracted trajectories from these estimations can sometimes be unrealistic when events happen in quick succession (although we note this issue is worse in the non-temporal models we compared our approach to). Future work could consider predicting the trajectory of an agent over a number of timesteps simultaneously to mitigate this issue.

Further studies could also compare our proposed \textit{Agent Imputer} model against state-of-the-art models that use tracking rather than event data (e.g., \cite{omidshafiei2022multiagent, le2017coordinated}). In this work, we have only compared to other models that use event data. Comparison to tracking data models would facilitate better comparison of the effectiveness of our imputation approach in downstream analysis tasks.

Finally, we used a relatively small number of games in this work (34 matches in total) as it is difficult to obtain tracking data. Future work could extend our dataset to include a larger number of games. This would help the model learn a wider range of game scenarios beyond the ones that occur within our current dataset. Furthermore, this would facilitate the development of more specific models, for example, focusing on a particular player or team.

\vspace{-0.1cm}
\section{Conclusion} \label{conc}

This paper presents a novel \textit{Agent Imputer} model to address a multi-agent behavioural prediction problem in dynamic environments with limited system observability. We apply this task to football by imputing off-ball player locations using only on-ball data. We find that our model can impute player location to within $\sim$6.9 metres, outperforming multiple baseline imputation models. We perform deeper analysis of model performance, examining accuracy over different game times, player roles, and observation rates. We also present football analytics applications facilitated by our model, which allow lower league clubs to perform similar processes to elite level clubs without requiring access to expensive data. This novel work could be applied to other real world domains involving limited agent visibility, such as disaster response or tracking daily human movement from phone data.

\newpage

\begin{acks}
We thank Bepro Group Ltd for supporting and providing the data resources for this work. Gregory Everett was supported by Sentient Sports and Sarvapali Ramchurn was supported by the UK Engineering and Physical Sciences Research Council (EPSRC) through the Trustworthy Autonomous Systems Hub (EP/V00784X/1) and the AXA Research Fund.
\end{acks}
\balance
\bibliographystyle{ACM-Reference-Format} 
\bibliography{sample}


\begin{thebibliography}{28}


\ifx \showCODEN    \undefined \def \showCODEN     #1{\unskip}     \fi
\ifx \showDOI      \undefined \def \showDOI       #1{#1}\fi
\ifx \showISBNx    \undefined \def \showISBNx     #1{\unskip}     \fi
\ifx \showISBNxiii \undefined \def \showISBNxiii  #1{\unskip}     \fi
\ifx \showISSN     \undefined \def \showISSN      #1{\unskip}     \fi
\ifx \showLCCN     \undefined \def \showLCCN      #1{\unskip}     \fi
\ifx \shownote     \undefined \def \shownote      #1{#1}          \fi
\ifx \showarticletitle \undefined \def \showarticletitle #1{#1}   \fi
\ifx \showURL      \undefined \def \showURL       {\relax}        \fi
\providecommand\bibfield[2]{#2}
\providecommand\bibinfo[2]{#2}
\providecommand\natexlab[1]{#1}
\providecommand\showeprint[2][]{arXiv:#2}

\bibitem[\protect\citeauthoryear{Alahi, Goel, Ramanathan, Robicquet, Fei-Fei,
  and Savarese}{Alahi et~al\mbox{.}}{2016}]%
        {alahi2016social}
\bibfield{author}{\bibinfo{person}{Alexandre Alahi}, \bibinfo{person}{Kratarth
  Goel}, \bibinfo{person}{Vignesh Ramanathan}, \bibinfo{person}{Alexandre
  Robicquet}, \bibinfo{person}{Li Fei-Fei}, {and} \bibinfo{person}{Silvio
  Savarese}.} \bibinfo{year}{2016}\natexlab{}.
\newblock \showarticletitle{Social lstm: Human trajectory prediction in crowded
  spaces}. In \bibinfo{booktitle}{\emph{Proceedings of the IEEE conference on
  computer vision and pattern recognition}}. \bibinfo{pages}{961--971}.
\newblock


\bibitem[\protect\citeauthoryear{Barris and Button}{Barris and Button}{2008}]%
        {barris2008review}
\bibfield{author}{\bibinfo{person}{Sian Barris} {and} \bibinfo{person}{Chris
  Button}.} \bibinfo{year}{2008}\natexlab{}.
\newblock \showarticletitle{A review of vision-based motion analysis in sport}.
\newblock \bibinfo{journal}{\emph{Sports Medicine}} \bibinfo{volume}{38},
  \bibinfo{number}{12} (\bibinfo{year}{2008}), \bibinfo{pages}{1025--1043}.
\newblock


\bibitem[\protect\citeauthoryear{Baytas, Xiao, Zhang, Wang, Jain, and
  Zhou}{Baytas et~al\mbox{.}}{2017}]%
        {baytas2017patient}
\bibfield{author}{\bibinfo{person}{Inci~M Baytas}, \bibinfo{person}{Cao Xiao},
  \bibinfo{person}{Xi Zhang}, \bibinfo{person}{Fei Wang},
  \bibinfo{person}{Anil~K Jain}, {and} \bibinfo{person}{Jiayu Zhou}.}
  \bibinfo{year}{2017}\natexlab{}.
\newblock \showarticletitle{Patient subtyping via time-aware LSTM networks}. In
  \bibinfo{booktitle}{\emph{Proceedings of the 23rd ACM SIGKDD international
  conference on knowledge discovery and data mining}}. \bibinfo{pages}{65--74}.
\newblock


\bibitem[\protect\citeauthoryear{Beal, Chalkiadakis, Norman, and Ramchurn}{Beal
  et~al\mbox{.}}{2020a}]%
        {beal2020optimising}
\bibfield{author}{\bibinfo{person}{Ryan Beal}, \bibinfo{person}{Georgios
  Chalkiadakis}, \bibinfo{person}{Timothy~J Norman}, {and}
  \bibinfo{person}{Sarvapali~D Ramchurn}.} \bibinfo{year}{2020}\natexlab{a}.
\newblock \showarticletitle{Optimising Game Tactics for Football}. In
  \bibinfo{booktitle}{\emph{Proceedings of the 19th International Conference on
  Autonomous Agents and MultiAgent Systems}}. \bibinfo{pages}{141--149}.
\newblock


\bibitem[\protect\citeauthoryear{Beal, Chalkiadakis, Norman, and Ramchurn}{Beal
  et~al\mbox{.}}{2021}]%
        {beal2021optimising}
\bibfield{author}{\bibinfo{person}{Ryan Beal}, \bibinfo{person}{Georgios
  Chalkiadakis}, \bibinfo{person}{Timothy~J Norman}, {and}
  \bibinfo{person}{Sarvapali~D Ramchurn}.} \bibinfo{year}{2021}\natexlab{}.
\newblock \showarticletitle{Optimising Long-Term Outcomes using Real-World
  Fluent Objectives: An Application to Football}. In
  \bibinfo{booktitle}{\emph{Proceedings of the 20th International Conference on
  Autonomous Agents and MultiAgent Systems}}. \bibinfo{pages}{196--204}.
\newblock


\bibitem[\protect\citeauthoryear{Beal, Changder, Norman, and Ramchurn}{Beal
  et~al\mbox{.}}{2020b}]%
        {beal2020learning}
\bibfield{author}{\bibinfo{person}{Ryan Beal}, \bibinfo{person}{Narayan
  Changder}, \bibinfo{person}{Timothy Norman}, {and} \bibinfo{person}{Sarvapali
  Ramchurn}.} \bibinfo{year}{2020}\natexlab{b}.
\newblock \showarticletitle{Learning the value of teamwork to form efficient
  teams}. In \bibinfo{booktitle}{\emph{Proceedings of the AAAI Conference on
  Artificial Intelligence}}, Vol.~\bibinfo{volume}{34}.
  \bibinfo{pages}{7063--7070}.
\newblock


\bibitem[\protect\citeauthoryear{Fern{\'a}ndez, Bornn, and
  Cervone}{Fern{\'a}ndez et~al\mbox{.}}{2019}]%
        {fernandez_decomposing_2019}
\bibfield{author}{\bibinfo{person}{Javier Fern{\'a}ndez}, \bibinfo{person}{Luke
  Bornn}, {and} \bibinfo{person}{Dan Cervone}.}
  \bibinfo{year}{2019}\natexlab{}.
\newblock \showarticletitle{Decomposing the immeasurable sport: A deep learning
  expected possession value framework for soccer}. In
  \bibinfo{booktitle}{\emph{13th MIT Sloan Sports Analytics Conference}}.
\newblock


\bibitem[\protect\citeauthoryear{Hamilton, Ying, and Leskovec}{Hamilton
  et~al\mbox{.}}{2017}]%
        {hamilton2017inductive}
\bibfield{author}{\bibinfo{person}{Will Hamilton}, \bibinfo{person}{Zhitao
  Ying}, {and} \bibinfo{person}{Jure Leskovec}.}
  \bibinfo{year}{2017}\natexlab{}.
\newblock \showarticletitle{Inductive representation learning on large graphs}.
\newblock \bibinfo{journal}{\emph{Advances in neural information processing
  systems}}  \bibinfo{volume}{30} (\bibinfo{year}{2017}).
\newblock


\bibitem[\protect\citeauthoryear{Hauri, Djuric, Radosavljevic, and
  Vucetic}{Hauri et~al\mbox{.}}{2021}]%
        {hauri2021multi}
\bibfield{author}{\bibinfo{person}{Sandro Hauri}, \bibinfo{person}{Nemanja
  Djuric}, \bibinfo{person}{Vladan Radosavljevic}, {and}
  \bibinfo{person}{Slobodan Vucetic}.} \bibinfo{year}{2021}\natexlab{}.
\newblock \showarticletitle{Multi-Modal Trajectory Prediction of NBA Players}.
  In \bibinfo{booktitle}{\emph{Proceedings of the IEEE/CVF Winter Conference on
  Applications of Computer Vision}}. \bibinfo{pages}{1640--1649}.
\newblock


\bibitem[\protect\citeauthoryear{Ivanovic and Pavone}{Ivanovic and
  Pavone}{2019}]%
        {ivanovic2019trajectron}
\bibfield{author}{\bibinfo{person}{Boris Ivanovic} {and} \bibinfo{person}{Marco
  Pavone}.} \bibinfo{year}{2019}\natexlab{}.
\newblock \showarticletitle{The trajectron: Probabilistic multi-agent
  trajectory modeling with dynamic spatiotemporal graphs}. In
  \bibinfo{booktitle}{\emph{Proceedings of the IEEE/CVF International
  Conference on Computer Vision}}. \bibinfo{pages}{2375--2384}.
\newblock


\bibitem[\protect\citeauthoryear{Le, Yue, Carr, and Lucey}{Le
  et~al\mbox{.}}{2017}]%
        {le2017coordinated}
\bibfield{author}{\bibinfo{person}{Hoang~M Le}, \bibinfo{person}{Yisong Yue},
  \bibinfo{person}{Peter Carr}, {and} \bibinfo{person}{Patrick Lucey}.}
  \bibinfo{year}{2017}\natexlab{}.
\newblock \showarticletitle{Coordinated multi-agent imitation learning}. In
  \bibinfo{booktitle}{\emph{International Conference on Machine Learning}}.
  PMLR, \bibinfo{pages}{1995--2003}.
\newblock


\bibitem[\protect\citeauthoryear{Link, Lang, and Seidenschwarz}{Link
  et~al\mbox{.}}{2016}]%
        {link_real_2016}
\bibfield{author}{\bibinfo{person}{Daniel Link}, \bibinfo{person}{Steffen
  Lang}, {and} \bibinfo{person}{Philipp Seidenschwarz}.}
  \bibinfo{year}{2016}\natexlab{}.
\newblock \showarticletitle{Real time quantification of dangerousity in
  football using spatiotemporal tracking data}.
\newblock \bibinfo{journal}{\emph{PloS one}} \bibinfo{volume}{11},
  \bibinfo{number}{12} (\bibinfo{year}{2016}), \bibinfo{pages}{e0168768}.
\newblock


\bibitem[\protect\citeauthoryear{Loshchilov and Hutter}{Loshchilov and
  Hutter}{2018}]%
        {loshchilov2018decoupled}
\bibfield{author}{\bibinfo{person}{Ilya Loshchilov} {and}
  \bibinfo{person}{Frank Hutter}.} \bibinfo{year}{2018}\natexlab{}.
\newblock \showarticletitle{Decoupled Weight Decay Regularization}. In
  \bibinfo{booktitle}{\emph{International Conference on Learning
  Representations}}.
\newblock


\bibitem[\protect\citeauthoryear{Lucey, Bialkowski, Carr, Foote, and
  Matthews}{Lucey et~al\mbox{.}}{2012}]%
        {lucey2012characterizing}
\bibfield{author}{\bibinfo{person}{Patrick Lucey}, \bibinfo{person}{Alina
  Bialkowski}, \bibinfo{person}{Peter Carr}, \bibinfo{person}{Eric Foote},
  {and} \bibinfo{person}{Iain Matthews}.} \bibinfo{year}{2012}\natexlab{}.
\newblock \showarticletitle{Characterizing multi-agent team behavior from
  partial team tracings: Evidence from the english premier league}. In
  \bibinfo{booktitle}{\emph{Proceedings of the AAAI Conference on Artificial
  Intelligence}}, Vol.~\bibinfo{volume}{26}. \bibinfo{pages}{1387--1393}.
\newblock


\bibitem[\protect\citeauthoryear{Marchetti, Becattini, Seidenari, and
  Del~Bimbo}{Marchetti et~al\mbox{.}}{2020}]%
        {marchetti2020multiple}
\bibfield{author}{\bibinfo{person}{Francesco Marchetti},
  \bibinfo{person}{Federico Becattini}, \bibinfo{person}{Lorenzo Seidenari},
  {and} \bibinfo{person}{Alberto Del~Bimbo}.} \bibinfo{year}{2020}\natexlab{}.
\newblock \showarticletitle{Multiple trajectory prediction of moving agents
  with memory augmented networks}.
\newblock \bibinfo{journal}{\emph{IEEE Transactions on Pattern Analysis and
  Machine Intelligence}} (\bibinfo{year}{2020}).
\newblock


\bibitem[\protect\citeauthoryear{McInerney, Rogers, and Jennings}{McInerney
  et~al\mbox{.}}{2012}]%
        {mcinerney2012improving}
\bibfield{author}{\bibinfo{person}{James McInerney}, \bibinfo{person}{Alex
  Rogers}, {and} \bibinfo{person}{Nicholas~R Jennings}.}
  \bibinfo{year}{2012}\natexlab{}.
\newblock \showarticletitle{Improving location prediction services for new
  users with probabilistic latent semantic analysis}. In
  \bibinfo{booktitle}{\emph{Proceedings of the 2012 ACM conference on
  ubiquitous computing}}. \bibinfo{pages}{906--910}.
\newblock


\bibitem[\protect\citeauthoryear{McInerney, Zheng, Rogers, and
  Jennings}{McInerney et~al\mbox{.}}{2013}]%
        {mcinerney2013modelling}
\bibfield{author}{\bibinfo{person}{James McInerney},
  \bibinfo{person}{Jiangchuan Zheng}, \bibinfo{person}{Alex Rogers}, {and}
  \bibinfo{person}{Nicholas~R Jennings}.} \bibinfo{year}{2013}\natexlab{}.
\newblock \showarticletitle{Modelling heterogeneous location habits in human
  populations for location prediction under data sparsity}. In
  \bibinfo{booktitle}{\emph{Proceedings of the 2013 ACM international joint
  conference on Pervasive and ubiquitous computing}}.
  \bibinfo{pages}{469--478}.
\newblock


\bibitem[\protect\citeauthoryear{Omidshafiei, Hennes, Garnelo, Wang, Recasens,
  Tarassov, Yang, Elie, Connor, Muller, et~al\mbox{.}}{Omidshafiei
  et~al\mbox{.}}{2022}]%
        {omidshafiei2022multiagent}
\bibfield{author}{\bibinfo{person}{Shayegan Omidshafiei},
  \bibinfo{person}{Daniel Hennes}, \bibinfo{person}{Marta Garnelo},
  \bibinfo{person}{Zhe Wang}, \bibinfo{person}{Adria Recasens},
  \bibinfo{person}{Eugene Tarassov}, \bibinfo{person}{Yi Yang},
  \bibinfo{person}{Romuald Elie}, \bibinfo{person}{Jerome~T Connor},
  \bibinfo{person}{Paul Muller}, {et~al\mbox{.}}}
  \bibinfo{year}{2022}\natexlab{}.
\newblock \showarticletitle{Multiagent off-screen behavior prediction in
  football}.
\newblock \bibinfo{journal}{\emph{Scientific reports}} \bibinfo{volume}{12},
  \bibinfo{number}{1} (\bibinfo{year}{2022}), \bibinfo{pages}{1--13}.
\newblock


\bibitem[\protect\citeauthoryear{Pappalardo, Cintia, Rossi, Massucco,
  Ferragina, Pedreschi, and Giannotti}{Pappalardo et~al\mbox{.}}{2019}]%
        {pappalardo2019public}
\bibfield{author}{\bibinfo{person}{Luca Pappalardo}, \bibinfo{person}{Paolo
  Cintia}, \bibinfo{person}{Alessio Rossi}, \bibinfo{person}{Emanuele
  Massucco}, \bibinfo{person}{Paolo Ferragina}, \bibinfo{person}{Dino
  Pedreschi}, {and} \bibinfo{person}{Fosca Giannotti}.}
  \bibinfo{year}{2019}\natexlab{}.
\newblock \showarticletitle{A public data set of spatio-temporal match events
  in soccer competitions}.
\newblock \bibinfo{journal}{\emph{Scientific data}} \bibinfo{volume}{6},
  \bibinfo{number}{1} (\bibinfo{year}{2019}), \bibinfo{pages}{1--15}.
\newblock


\bibitem[\protect\citeauthoryear{Qi, Qin, Wu, and Yang}{Qi
  et~al\mbox{.}}{2020}]%
        {qi2020imitative}
\bibfield{author}{\bibinfo{person}{Mengshi Qi}, \bibinfo{person}{Jie Qin},
  \bibinfo{person}{Yu Wu}, {and} \bibinfo{person}{Yi Yang}.}
  \bibinfo{year}{2020}\natexlab{}.
\newblock \showarticletitle{Imitative non-autoregressive modeling for
  trajectory forecasting and imputation}. In
  \bibinfo{booktitle}{\emph{Proceedings of the IEEE/CVF Conference on Computer
  Vision and Pattern Recognition}}. \bibinfo{pages}{12736--12745}.
\newblock


\bibitem[\protect\citeauthoryear{Raabe, Nabben, and Memmert}{Raabe
  et~al\mbox{.}}{2022}]%
        {raabe2022graph}
\bibfield{author}{\bibinfo{person}{Dominik Raabe}, \bibinfo{person}{Reinhard
  Nabben}, {and} \bibinfo{person}{Daniel Memmert}.}
  \bibinfo{year}{2022}\natexlab{}.
\newblock \showarticletitle{Graph representations for the analysis of
  multi-agent spatiotemporal sports data}.
\newblock \bibinfo{journal}{\emph{Applied Intelligence}}
  (\bibinfo{year}{2022}), \bibinfo{pages}{1--21}.
\newblock


\bibitem[\protect\citeauthoryear{Ramchurn, Huynh, Wu, Ikuno, Flann, Moreau,
  Fischer, Jiang, Rodden, Simpson, et~al\mbox{.}}{Ramchurn
  et~al\mbox{.}}{2016}]%
        {ramchurn2016disaster}
\bibfield{author}{\bibinfo{person}{Sarvapali~D Ramchurn},
  \bibinfo{person}{Trung~Dong Huynh}, \bibinfo{person}{Feng Wu},
  \bibinfo{person}{Yukki Ikuno}, \bibinfo{person}{Jack Flann},
  \bibinfo{person}{Luc Moreau}, \bibinfo{person}{Joel~E Fischer},
  \bibinfo{person}{Wenchao Jiang}, \bibinfo{person}{Tom Rodden},
  \bibinfo{person}{Edwin Simpson}, {et~al\mbox{.}}}
  \bibinfo{year}{2016}\natexlab{}.
\newblock \showarticletitle{A disaster response system based on human-agent
  collectives}.
\newblock \bibinfo{journal}{\emph{Journal of Artificial Intelligence Research}}
   \bibinfo{volume}{57} (\bibinfo{year}{2016}), \bibinfo{pages}{661--708}.
\newblock


\bibitem[\protect\citeauthoryear{Spearman}{Spearman}{2018}]%
        {spearman_beyond_2018}
\bibfield{author}{\bibinfo{person}{William Spearman}.}
  \bibinfo{year}{2018}\natexlab{}.
\newblock \showarticletitle{Beyond expected goals}. In
  \bibinfo{booktitle}{\emph{Proceedings of the 12th MIT sloan sports analytics
  conference}}. \bibinfo{pages}{1--17}.
\newblock


\bibitem[\protect\citeauthoryear{Spearman, Basye, Dick, Hotovy, and
  Pop}{Spearman et~al\mbox{.}}{2017}]%
        {spearman2017physics}
\bibfield{author}{\bibinfo{person}{William Spearman}, \bibinfo{person}{Austin
  Basye}, \bibinfo{person}{Greg Dick}, \bibinfo{person}{Ryan Hotovy}, {and}
  \bibinfo{person}{Paul Pop}.} \bibinfo{year}{2017}\natexlab{}.
\newblock \showarticletitle{Physics-based modeling of pass probabilities in
  soccer}. In \bibinfo{booktitle}{\emph{Proceeding of the 11th MIT Sloan Sports
  Analytics Conference}}.
\newblock


\bibitem[\protect\citeauthoryear{Sriram, Liu, Pittaluga, and Chandraker}{Sriram
  et~al\mbox{.}}{2020}]%
        {sriram2020smart}
\bibfield{author}{\bibinfo{person}{NN Sriram}, \bibinfo{person}{Buyu Liu},
  \bibinfo{person}{Francesco Pittaluga}, {and} \bibinfo{person}{Manmohan
  Chandraker}.} \bibinfo{year}{2020}\natexlab{}.
\newblock \showarticletitle{Smart: Simultaneous multi-agent recurrent
  trajectory prediction}. In \bibinfo{booktitle}{\emph{European Conference on
  Computer Vision}}. Springer, \bibinfo{pages}{463--479}.
\newblock


\bibitem[\protect\citeauthoryear{Sun, Karlsson, Wu, Tenenbaum, and Murphy}{Sun
  et~al\mbox{.}}{2018}]%
        {sun2019stochastic}
\bibfield{author}{\bibinfo{person}{Chen Sun}, \bibinfo{person}{Per Karlsson},
  \bibinfo{person}{Jiajun Wu}, \bibinfo{person}{Joshua~B Tenenbaum}, {and}
  \bibinfo{person}{Kevin Murphy}.} \bibinfo{year}{2018}\natexlab{}.
\newblock \showarticletitle{Stochastic Prediction of Multi-Agent Interactions
  from Partial Observations}. In \bibinfo{booktitle}{\emph{International
  Conference on Learning Representations}}.
\newblock


\bibitem[\protect\citeauthoryear{Wang, Cai, Yue, and Suresh}{Wang
  et~al\mbox{.}}{2021}]%
        {wang2021pre}
\bibfield{author}{\bibinfo{person}{Jing Wang}, \bibinfo{person}{Jianping Cai},
  \bibinfo{person}{Xiaohang Yue}, {and} \bibinfo{person}{Nallan~C Suresh}.}
  \bibinfo{year}{2021}\natexlab{}.
\newblock \showarticletitle{Pre-positioning and real-time disaster response
  operations: Optimization with mobile phone location data}.
\newblock \bibinfo{journal}{\emph{Transportation research part E: logistics and
  transportation review}}  \bibinfo{volume}{150} (\bibinfo{year}{2021}),
  \bibinfo{pages}{102344}.
\newblock


\bibitem[\protect\citeauthoryear{Xie, Zhang, Zhu, Wu, and Zhu}{Xie
  et~al\mbox{.}}{2021}]%
        {xie2021congestion}
\bibfield{author}{\bibinfo{person}{Xu Xie}, \bibinfo{person}{Chi Zhang},
  \bibinfo{person}{Yixin Zhu}, \bibinfo{person}{Ying~Nian Wu}, {and}
  \bibinfo{person}{Song-Chun Zhu}.} \bibinfo{year}{2021}\natexlab{}.
\newblock \showarticletitle{Congestion-aware multi-agent trajectory prediction
  for collision avoidance}. In \bibinfo{booktitle}{\emph{2021 IEEE
  International Conference on Robotics and Automation (ICRA)}}. IEEE,
  \bibinfo{pages}{13693--13700}.
\newblock


\end{thebibliography}

\newpage
\appendix

\setcounter{table}{0}
\renewcommand{\thetable}{A\arabic{table}}

\setcounter{figure}{0}
\renewcommand{\thefigure}{A\arabic{figure}}

\onecolumn

\section{Implementation Details}
\label{app:implementation}

In this section, we list the compute resources used in this work (Section \ref{app:compute}), and provide further details on the feature embedding process (Section \ref{app:encoding}).

\subsection{Compute Resources}
\label{app:compute}

This work was implemented in Python 3.8.8 and our machine learning functionality used PyTorch.\footnote{https://pytorch.org/} Model training was carried out on a remote GPU service using a NVIDIA V100 GPU with 32 GB of VRAM facilitated by Google Colab. This used CUDA to enable training on the GPU. Training the \textit{Agent Imputer} model took a maximum of three hours for each fold. Code is accesible in a public Github repository found here: https://github.com/GregSoton/PlayerImputation.

\subsection{Feature Embedding}
\label{app:encoding}

We embedded our categorical variables (\texttt{\small{agentRole}}, \texttt{\small{agentSide}}, \texttt{\small{agentObserved}}, \texttt{\small{goalDiff}}, \texttt{\small{eventType}}) using random embedding vectors where each vector had size equal to the square root of the number of classes for each variable. This increased the total categorical feature dimensionality from 5 to 14. Future work should consider the use of trainable embeddings. We normalise spatial features using individual min-max scalers for the X and Y directions. For time-based features, we use the RobustScaler from the scikit-learn package.\footnote{https://scikit-learn.org/stable/index.html.} This appropriately scales time whilst being robust to outliers caused by abnormally long periods between events (e.g., an injury).

\section{Dataset Details}
\label{app:additional}

In this section we provide further details on the dataset structure (Section \ref{app:dataset}) and player positions (Section \ref{app:positions})

\subsection{Dataset Structure}
\label{app:dataset}

The ground truth tracking dataset we use logs player position at 30 Hz. To align player tracking with event data, we find the closest tracking timestamp to the event timestamp (usually within a few milliseconds). For our data, the set of possible event types are: Pass, Pass Received, Foul, Foul Won, Duel, Tackle, Shot, Save, Intervention, Recovery, Interception, Take-On, Block, Ball Received, Clearance, Error, Goal Conceded, Offside, Aerial Clearance, Pause, Own Goal, Control Under Pressure, Deflection, Carry, Substitution. The player who receives a pass is not highlighted for a pass event, and a pass received event is instead referenced. This pattern is consistent throughout the data, whereby related events are referenced for each event.

\subsection{Player Positions}
\label{app:positions}

The event data provided gives a label for player role in each game. There are 16 unique roles which we pass into our model. When applying downstream analysis, we group roles for clarity and to increase the amount of data for each role. Below, we present the role groups (bold) and roles (nested bullet points):

\begin{multicols}{2}
    \begin{itemize}
        \item \textbf{Goalkeeper}
        \begin{itemize}
            \item Goalkeeper
        \end{itemize}
        \item \textbf{Central Defender}
        \begin{itemize}
            \item Center Back
            \item Central Defensive Midfielder
        \end{itemize}
        \item \textbf{Wide Defender}
        \begin{itemize}
            \item Left Back
            \item Right Back
            \item Left Wing Back
            \item Right Wing Back
        \end{itemize}
        \item \textbf{Central Midfielder}
        \begin{itemize}
            \item Central Midfielder
        \end{itemize}
        \item \textbf{Wide Midfielder}
        \begin{itemize}
            \item Left Midfielder
            \item Right Midfielder
        \end{itemize}
        \item \textbf{Central Attacker}
        \begin{itemize}
            \item Central Attacking Midfielder
            \item Centre Forward
        \end{itemize}
        \item \textbf{Wide Attacker}
        \begin{itemize}
            \item Left Winger
            \item Right Winger
            \item Left Forward
            \item Right Forward
        \end{itemize}
   \end{itemize}
\end{multicols}
\clearpage

\section{Additional Player Position Heatmaps}
\label{app:heatmaps}

\begin{figure*}[!h]
  \includegraphics[width=\textwidth]{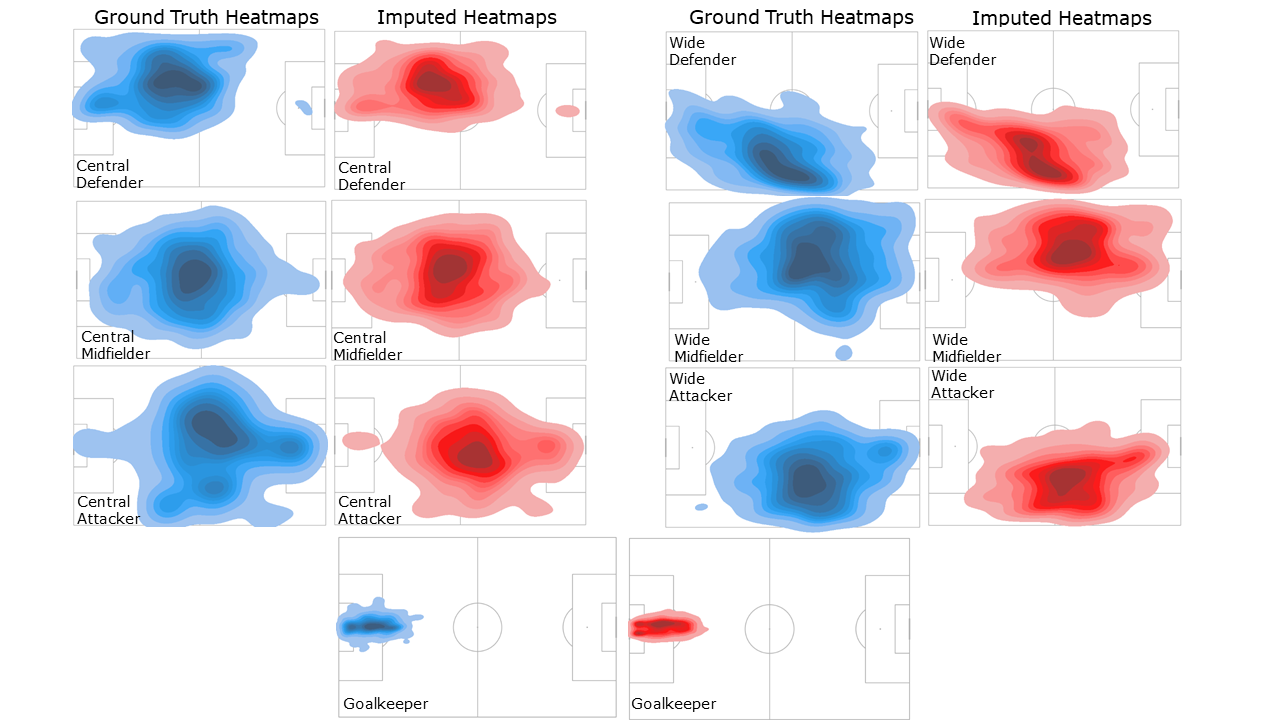}
  \vspace{-0.6cm}
  \caption{Player heatmaps for all roles.}
  \label{fig:all_heatmaps}
  \vspace{-0.3cm}
\end{figure*}

In this section, we provide additional player heatmaps for each role; see Figure \ref{fig:all_heatmaps}.

\end{document}